\documentclass[11pt]{article}
\usepackage{acl2016}
\usepackage{times}
\usepackage{latexsym}
\usepackage{graphicx}
\usepackage{amsmath}
\usepackage{amssymb}
\usepackage{multirow}
\usepackage[font={sf,bf},textfont=md]{caption}
\usepackage{CJK}
\begin{CJK}{UTF8}{gbsn}
\aclfinalcopy % Uncomment this line for the final submission
\title{Modeling Rich Contexts for Sentiment Classification with LSTM}

\author{Minlie Huang, Yujie Cao, Chao Dong\\
	    State Key Lab. of Intelligent Technology and Systems, National Lab. for\\ Information Science and Technology, Dept. of Computer Science and Technology,\\ Tsinghua University, Beijing 100084, PR China\\
	    {\tt aihuang@tsinghua.edu.cn, caoyujieboy@163.com, neutronest@gmail.com}	    
}

\date{}

\begin{document}

\maketitle

\begin{abstract}
Sentiment analysis on social media data such as tweets and weibo has become a very important and challenging task. Due to the intrinsic properties of such data, tweets are short, noisy, and of divergent topics, and sentiment classification on these data requires to modeling various contexts such as the retweet/reply history of a tweet, and the social context about authors and relationships. While few prior study has approached the issue of modeling contexts in tweet, this paper proposes to use a hierarchical LSTM to model rich contexts in tweet, particularly long-range context. Experimental results show that contexts can help us to perform sentiment classification remarkably better.
\end{abstract}

\section{Introduction}
Social media have been a major source for people to express their opinions online. Sentiment analysis on social media like Twitter and Weibo has drawn more and more attentions recently. However, due to the intrinsic properties of social media data, tweet is short, noisy, and of divergent topics. Sentiment analysis on such data requires to model contexts for a current tweet, for instance, to take into account social context (such as the relations between follower and followee), discourse relations (such as the connectives and conditionals between sentences), topic-based and conversation-based context, and many more. It is even more challenging to consider long-range contexts such as the entire retweet/reply history for a tweet, which has been largely ignored by prior studies.

\begin{figure}

\centering
\includegraphics[width=\columnwidth,clip=true]{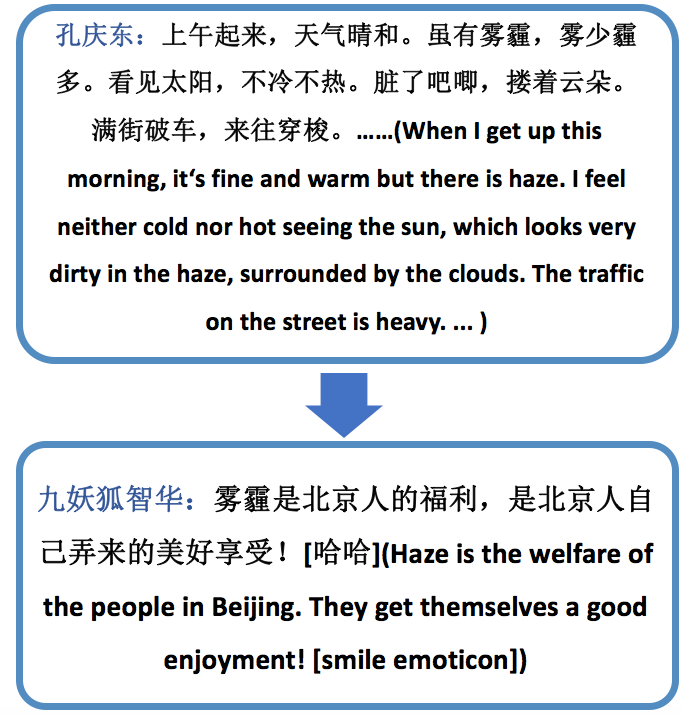}
\caption{An example tweet thread.}
\label{fig1:example}
\end{figure}

The long-range context indeed influences sentiment classification of a tweet. By analysing a dataset consisting of more than 14,000 tweets in about 1,600 threads where each thread consists of a sequence of retweeting/replying tweets discussing the same event or topic, we find out that 46.9\% of tweets have the same sentiment polarities with their root tweets in the re-tweeting process and 64.7\% with their parent tweets (three-class polarity: positive, neutral, and negative). That means, the polarity of an origin tweet has drifted to a large degree during the retweeting/replying process while neighboring tweets have higher possibility to maintain the same polarity. %With the long-range context we can have a more comprehensive understanding of a targeted tweet. 
See the example in Figure \ref{fig1:example} which shows an origin tweet and a reply tweet. The sentiment polarity of the reply tweet is easy to be misclassified if we only predict it based on the its own content, since it contains several positive words such as ``福利 (welfare)'' and ``享受 (enjoyment)''. If we trace back to its ``ancestor'' tweet, we can judge the polarity correctly.

As a matter of fact, many works have utilized some kinds of contextual information of tweets. Deng et al.~\shortcite{Deng:2013} and Hu et al.~\shortcite{Hu:2013-1} studied to use social relationships of authors of tweets, and they pointed out that authors with closer social relationships tend to show more similar sentiment polarities towards a target topic. Mukherjee and Bhattacharyya~\shortcite{Mukherjee:2012} took advantage of discourse relationships between sentences as feature, such as connectives, conditionals and semantic operators etc. Vanzo et al.~\shortcite{Vanzo:2014} model a sequence of tweets in a conversation or containing same topics with $SVM^{HMM}$. Recently, Ren et al.~\shortcite{Ren:2016} proposed a context-sensitive neural network in Twitter sentiment analysis. They use three kinds of contextual information, such as tweets in a conversation, those with same author and sharing same hashtags. Ghosh et al.~\shortcite{Ghosh:2016} proposed contextual LSTM by incorporating contextual features like topics of preceeding and current sentences into the model. However, all these studies haven't fully utilized the long-range contextual information.

In this work we propose a hierarchical LSTM model with two levels of LSTM networks to model the retweeting/replying process and capture the long-range dependency. The first LSTM is a word-level LSTM, which can generate a representation of a single tweet. And the second is a tweet-level LSTM which can model the long-range context of the current tweet. Moreover, we incorporate some additional context features into our model, which will be shown in detail later.

To evaluate how the proposed model performs on sentiment classification, particularly with long-range contexts, we collect a new dataset from Weibo.com. For each topic, we collect tens of threads where each thread consists of a sequence of retweeting/replying tweets to the origin tweet. Different from previous studies that usually define contextual tweets as those containing the same hashtags, we believe that retweet/reply is more focused on topic and thus context can play a more important role in sentiment classification.

To sum up, the main contribution of this paper is as follows:
\begin{itemize}
\item We collect a new dataset from Weibo.com, where there are about 50 topics, 1,600 threads, and 14,000 tweets. Each tread consists of a sequence of retweeting/replying tweets.

\item We propose a hierarchical LSTM model containing two level LSTM networks. This model can capture long-range dependencies between a tweet and its contextual tweets.

\item We adopt additional contexts including social context and text-based context to help the model capture more features, and these contexts are derived from the tweets automatically.

%\item Experiment results show that our model outperforms the baselines remarkably. 

\end{itemize}

The rest of the paper is organized as follows. In Section 2, we will overview some related works. In Section 3, we will describe our proposed methods in detail, including our hierarchical LSTM model, some context features and the way how we add them. And we present the experiments in Section 4. We conclude this work in Section 5.

\section{Related Work}
\subsection{Twitter Sentiment Analysis}
Twitter sentiment analysis has been a very popular topic in the field of NLP these years. Many existing works were based on bag-of-words representation. Following Pang et al.~\shortcite{Pang:2002}, many works attempted to design more effective features such as emotional signals, lexicons, n-grams and so on~\cite{Hu:2013,Pierpaolo:2015,Taboada:2011,Feldman:2013}. Mohammad et al.~\shortcite{Mohammad:2013} built their system in SemEval-2013 with a number of  features like POS tags, hashtags, characters in upper case, punctuations and so on. 

Recently, many methods leveraged contextual information. Based on the characteristics of social network, there are social relationships that can be utilized to enhance the performance of predicting the polarities of tweets~\cite{Tan:2011,Deng:2013,Hu:2013-1}. Speriosu et al.~\shortcite{Speriosu:2011} proposed a label propagation algorithm by combining textual features as well as social context. Another important kind of context for sentiment analysis is the discourse relations between sentences~\cite{Asher:2008}. Somasundaran et al.~\shortcite{Somasundaran:2008} proposed \emph{``opinion frame''} as a representation to capture the discourse features. Based on this frame, Somasundaran et al.~\shortcite{Somasundaran:2009} modelled the discourse relations as graph, and conducted sentiment classification on the graph by means of collective classification. Yang and Cardie~\shortcite{Yang:2014} modelled the discourse relations as constraints with the posterior regularization learning framework. For Twitter sentiment analysis, Mukherjee and Bhattacharyya~\shortcite{Mukherjee:2012} utilized the discourse relations like conjunctions and conditionals between sentences as context. There are also other kinds of contextual information like topic-based and conversation-based contexts. Vanzo et al.~\shortcite{Vanzo:2014} proposed a context-based $SVM^{HMM}$ model which models a sequence of tweets in a conversation or in a same topic. 

A more recent work is the context-sensitive neural network \cite{Ren:2016} where embeddings of keywords extracted from contexts are pooled to the output layer together with the CNN output layer. An extension of LSTM named contextual LSTM was presented by Ghosh et al.~\shortcite{Ghosh:2016}, incorporating contextual features like topics of the preceeding, current sentences or paragraphs.

In comparison, our model can not only make use of some ordinary context features, but also model the long-range context features of tweets during the retweet/reply process by taking advantage of the hierarchical LSTM.
\subsection{Deep Learning Approaches for Sentiment Analysis}

Deep learning approaches have also benefited a variety of sentiment analysis tasks. One class of the deep learning approaches is convolutional neural network. Many works have built CNN architectures to model semantic and sentiment information of sentences ~\cite{Kalchbrenner:2014,Kim:2014}. In Semeval-2015, Severyn et al.~\shortcite{Severyn:2015} obtained top performance on predicting polarities at the message and phrase level by a deep CNN.

Another class of the deep learning approaches for sentiment analysis is recursive neural network introduced by Socher et al.~\shortcite{Socher:2011,Socher:2013} which composes a phrase recursively from its child phrases. Dong et al.~\shortcite{Dong:2014} proposed an adaptive RNN which combines a set of composition functions whose weights are learned adaptively. A deep RNN which stacks multiple recursive layers performed well in fine-grained sentiment analysis~\cite{Irsoy:2014}.

Some sequence models such as recurrent neural network, are suitable for sentiment classification tasks, since a sentence is composed of a sequence of words and similarly a document is composed of a sequence of sentences. Some recurrent neural networks were used to make prediction for words~\cite{Zhang:2014} or sequence labelling~\cite{Irsoy:2014-2}. Tai et al.~\shortcite{Tai:2015} proposed a tree-structured LSTM beating other sequential models in sentiment classification. Wang et al.~\shortcite{Wang:2015} cast a tweet as a sequence of words using LSTM to generate a sentence representation for predicting the polarities of tweets. Further Tang et al.~\shortcite{Tang:2015} utilized LSTM and gated RNN on the document level sentiment classification task.

\begin{figure*}

\centering
\includegraphics[width=\textwidth,clip=true]{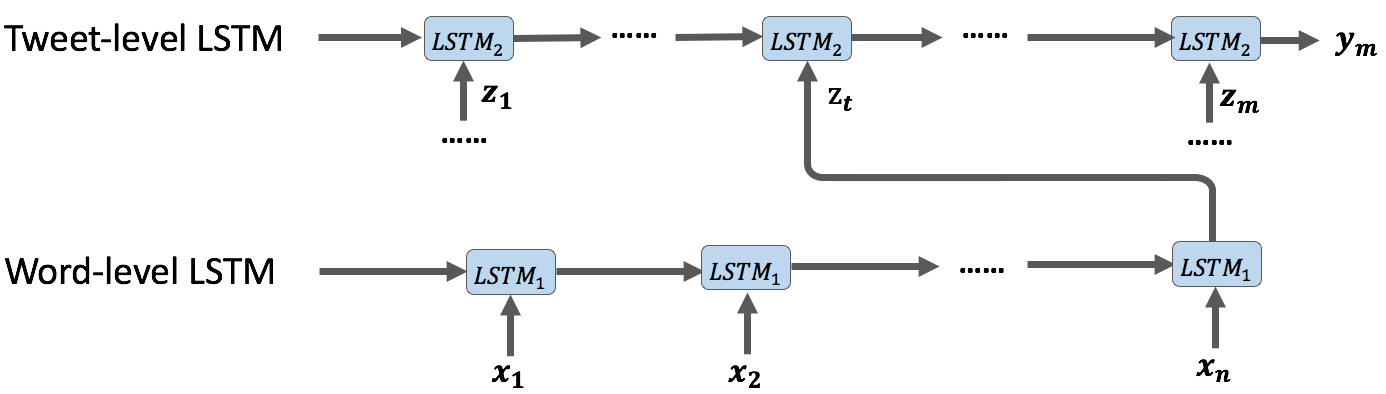}
\caption{Hierarchical Long Short-Term Memory (HLSTM)}

\end{figure*}

\section{Methodology}
\subsection{Long Short-Term Memory (LSTM)}

Some sequence models such as recurrent neural network are very suitable for sentiment analysis, since they can model the long-range dependency and thus utilize the contextual information. However, recurrent neural network has a vital problem of vanishing and exploding gradients~\cite{Bengio:1994}. Long Short-Term Memory (LSTM)~\cite{Hochreiter:1997} can tackle this issue to model long-range dependency. Like other recurrent neural networks, LSTM also has a recurrent layer consisting of memory blocks. In each memory block, there is a memory cell unit which can store memory state information and several gates which can control the change of the memory state. More formally, we have
\begin{align}
  & f_{t} = \sigma(W_{fz}z_{t-1} + W_{fx}x_{t}  + b_f) \\
  & i_{t} = \sigma(W_{iz}z_{t-1} + W_{ix}x_{t}  + b_i) \\
  & \tilde{C}_{t} = \phi(W_{cz}z_{t-1} + W_{cx}x_{t} + b_c) \\
  & C_{t} =  i_{t} \odot \tilde{C}_{t} + f_{t} \odot C_{t-1} \\
  & o_{t} = \sigma(W_{oz}z_{t-1} + W_{ox}x_{t} + b_o) \\
  & z_{t} = o_{t} \odot \phi(C_{t})
\end{align}

Here, $i_{t}$, $f_{t}$ and $o_{t}$ denote the input gate, forget gate and output gate, respectively. $x_t$ is an input vector and $z_t$ is the hidden representation. $W$ and $b$ are the weight matrix and the bias term respectively. $\sigma$ is $sigmoid$ and $\phi$ is $tanh$. $\odot$ is the element-wise multiplication.

As can be seen, the input, output and the cell state is controlled by the gates, and in this way the LSTM can decide to remember or forget the information in the recurrent layer. This gives this model the capacity of learning the long-term dependencies, which is very helpful for our task. For a single tweet, LSTM may focus more on the important words and generate a more meaningful representation for the tweet. Similarly, for a tweets thread, LSTM may focus more on the important tweets in the propagation process. Based on this idea, we propose our model, a two-level Hierarchical LSTM.

\subsection{Hierarchical LSTM (HLSTM) for Twitter Sentiment Analysis}

Since many tweets are short and informal in grammar, there is very limited information available in a single tweet. However, the tweets which have the relations of reply or retweet with the targeted tweet can be utilized as long-range context. This inspires us to design another layer of LSTM above the word-level LSTM.

From Figure 2 we can see that in the word-level LSTM, the input is individual words. The hidden state of the last word is taken as the representation of the tweet. Each tweet in a thread will go through the word-level LSTM, and the $t$-th tweet in the thread will generate a tweet representation of $z_{t}$ which will be an input of the tweet-level LSTM.

The inputs for the tweet-level LSTM is the representations of tweets in a propagation thread. The first one is the origin tweet, which is the base of the whole thread, while the others are the retweets or replies of the origin tweet. 
And the last one (here we suppose it's the $m$-th) is the targeted tweet that we are determined to predict the polarity. The output of the tweet-level LSTM at $m$ is $y_{m}$ which represents the $m$-th tweet with all the preceeding tweets. In this way the tweet-level LSTM incorporates the long-range context. We use the function of $softmax$ on $y_{m}$ to make sentiment classification. This model combines both the local features of a tweet and the long-range context to perform a context-aware analysis.

\subsection{HLSTM with Additional Contexts}

In addition to the long-range context, tweets have many additional contexts, such as social context, conversation-based context and topic-based context. These contextual information is very easy to obtain and they can be utilized properly to enhance our model. 

These additional contextual information is extracted from the parent tweet or the root tweet of a targeted tweet in the thread, which will be shown in detail in the next subsection. For each type of context, we take it as a binary-value feature and encode it into a $0$ or $1$. %If we take m kinds of contexts into consideration, we will get a $m$-digit binary vector $d$. 
These context features can be incorporated into the tweet-level LSTM, as follows:
\begin{align}
  & f_{t} = \sigma(W_{fh}h_{t-1} + W_{fz}z_{t} + W_{fd}d_{t} + b_f) \\
  & i_{t} = \sigma(W_{ih}h_{t-1} + W_{iz}z_{t} + W_{id}d_{t} + b_i) \\
  & \tilde{C}_{t} = \phi(W_{ch}h_{t-1} + W_{cz}z_{t}+ W_{cd}d_{t} + b_c) \\
  & C_{t} =  i_{t} \odot \tilde{C}_{t} + f_{t} \odot C_{t-1} \\
  & o_{t} = \sigma(W_{oh}h_{t-1} + W_{oz}z_{t} + W_{od}d_{t} + b_o) \\
  & h_{t} = o_{t} \odot \phi(C_{t})
\end{align}
where $d_t$ is the feature vector and $z_t$ is the input tweet representation from the first level LSTM.
From the above equations we can see that the additional context feature vector is fed into every gate. Actually, they can be regarded as an extension of the input vector $x$, which enrichs the representation of a targeted tweet. The way we add context features can be shown in Figure 3.

\begin{figure}
\centering
\includegraphics[width=\columnwidth,clip=true]{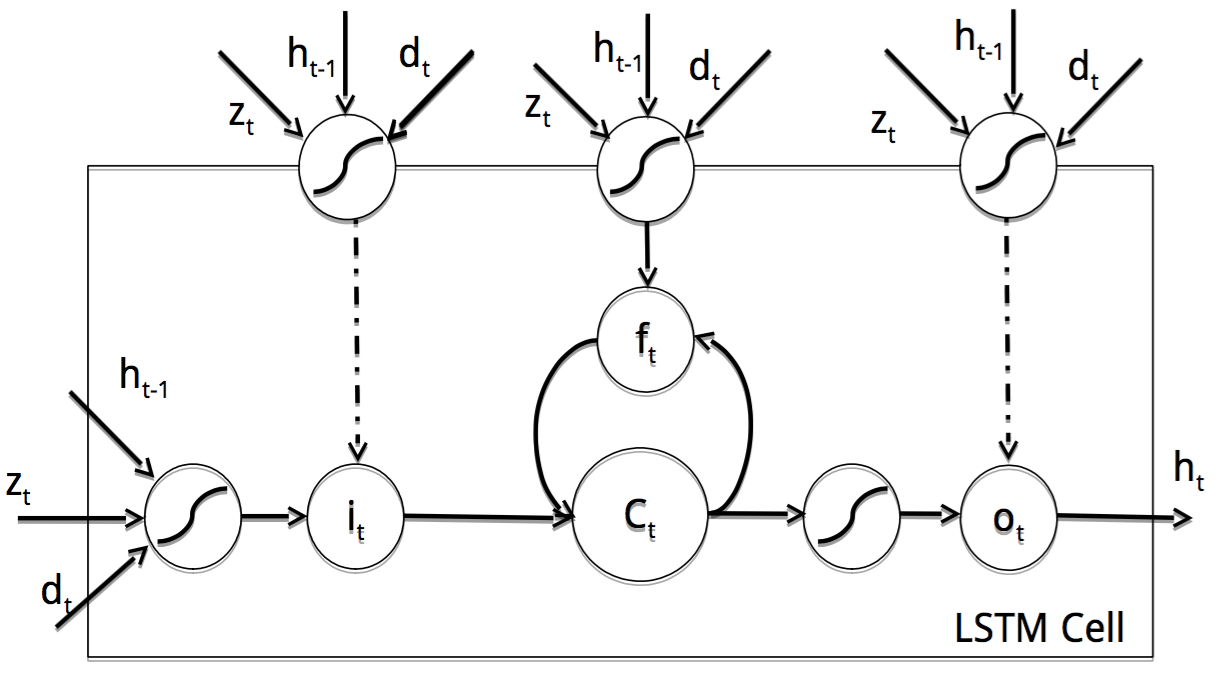}
\caption{LSTM cell of the tweet level LSTM network.}

\end{figure}

%$i_t$, $f_t$ and $o_t$ are the input gate, forget gate and output gate of time $t$ respectively. $C_t$ is the cell unit of the LSTM cell. $h_t$ is the hidden state of time t. $x_t$ is the input tweet representation from the first level LSTM and $d_t$ is the feature vector.  

\subsection{Context Features}

We take three types of contexts into account, the social context, the conversation-based context and the topic-based contexts. These contexts will be formed as binary features. We use tweet $j$ to denote the current tweet, and tweet $i$ to denote its parent tweet or its root tweet. 

The first one is a social context.

\textbf{SameAuthor} A same person usually holds consistent attitude towards a specific topic, especially in a tweet thread. If two tweets in the thread is posted by the same person, the two tweets tend to have same sentiment polarity. This context can be formulated as:
\begin{eqnarray}SameAuthor(i,j)=
\begin{cases}
1, &a_i=a_j\cr 0, &else
\end{cases}
\end{eqnarray}

$a_{i}$ and $a_{j}$ are the authors of tweet $i$ and tweet $j$ respectively. We should notice that tweet $i$ is the parent tweet or the root tweet, so that we get two features from the SameAuthor context for the current tweet. Similar for the following contexts.

The second one is a conversation-based context.

\textbf{Conversation} In a tweet thread, some tweets will mention other users' names by the function of $@username$. This indicates that the author of this tweet wants to have a conversation with or to show his opinions towards who has been mentioned. If a user who has been mentioned posts a tweet as response, then the two tweets constitute a conversation. The relation can be encoded as below:
\begin{eqnarray}Conversation(i,j)=
\begin{cases}
1, & a(j) \in M(i) \\ 0, &else
\end{cases}
\end{eqnarray}

Here, $a_{j}$ is the author of tweet $j$ and $M(i)$ is the set of users' names mentioned in tweet $i$.

The third type of contexts is topic-based contexts. We suppose that tweets which contain overlapping hashtags have similar topics; and tweets which contains overlapping emoticons have similar sentiment polarity towards their topics. 

\textbf{SameHashtag} In tweets, hashtags are used to highlight a topic. In a tweet thread, tweets on the same topic tend to show similar sentiment polarities.
\begin{eqnarray}SameHashtag(i,j)=
\begin{cases}
1, &H(i)\cap H(j)\neq\varnothing\\ 0, &else
\end{cases}
\end{eqnarray}

$H(i)$ and $H(j)$ are the sets of hashtags of tweet $i$ and tweet $j$.

\textbf{SameEmoji} Emoticons are used to express emotions. In a tweet thread, tweets with overlapping emoticons might have similar sentiment polarities towards their topics.
\begin{eqnarray}SameEmoji(i,j)=
\begin{cases}
1, &E(i)\cap E(j)\neq\varnothing\\ 0, &else
\end{cases}
\end{eqnarray}

$E(i)$ and $E(j)$ are the sets of emoticons of tweet $i$ and tweet $j$.

\subsection{Training}

The training objective of our model is the cross-entropy over the training examples. Here we use $P^g(x_i)$ to denote the golden-standard sentiment distribution of tweet $i$, and use $P(x_i)$ to denote the sentiment distribution that our model predicts. The loss function is shown below.
\begin{align}
  & loss = -\frac{1}{N}\sum_{i=1}^N\sum_{j=1}^C P^g_j(x_i)log(P_j(x_i))\\
  & P(x_i) = Softmax(h_i)
\end{align}

We use $N$ to denote the total number of training examples and $C$ the number of sentiment classes. Here $h_i$ is the hidden state of the Tweet level LSTM of the tweet $i$. 

We use AdaDelta~\cite{AdaDelta} to optimize the parameters.

\section{Experiments}
\subsection{Dataset}

We collected tweets from a Chinese Microblog website， Weibo.com. In Weibo.com, the retweets and replies always maintain their preceding tweets during the propagation process, and therefore long-range contexts are visible to a current tweet. We collected about 15k tweets consisting of over 1.6k tweet threads of 51 topics. We should notice that a tweet thread forms a tree structure started with an origin tweet, and the other nodes in the tree are retweets or replies. The average number of tweets in all the threads is 8.93, and the average depth is 3.75.

The sentiment polarity of each tweet is labelled by two independent annotators into three classes, $positive$, $neutral$ and $negative$. The annotation consistency of the dataset is 63.4\%, and those tweets with inconsistent labels are then checked by a third judger. Finally, there are   36\%, 39.6\%, and 24.6\% of positive, negative, and neutral tweets respectively. By analysing the dataset, we find that 46.9\% of tweets have the same sentiment polarities with their root tweets and 64.7\% with their parent twees , indicating that long-range context indeed influences the classification task. %This dataset has been used for an evaluation task on the COAE2015 conference\footnote{\tt http://www.ccir2015.com/nd.jsp?id=8}. 
The statistics of the dataset is summarized in Table 1.

\begin{table}
\centering
\small

\begin{tabular}{|l|l|}
\hline

\verb|#Topic| & {51} \\
\verb|#Thread| & {1649} \\
\verb|#Tweet| & {14733} \\ 
\verb|Average Thread Size| & {8.93} \\ 
\verb|Average Thread Depth| & {3.75} \\
\verb|%Same Sentiment with Root| & {46.9\%}  \\ 
\verb|%Same Sentiment with Parent| & {64.7\%}  \\
\verb|%Labelling Consistency| & {62.4\%}\\\hline
\end{tabular} 
\caption{Dataset statistics}
\end{table}

\subsection{Baseline Methods}

We compare our approach with the following baselines.
\begin{itemize}

\item SVM\(^{multiclass}\): a classical method for multi-class classification, which doesn't take any contextual information into account. In SVM\(^{multiclass}\), we use emoticons and bag-of-words as features, and train the classifier with SVM\(^{light}\).

\item SVM\(^{HMM}\) \cite{Vanzo:2014}: a SVM variant that combines HMM. This method follows the Markov first order hypothesis. In SVM\(^{HMM}\), we use the same features as those in SVM\(^{multiclass}\), and we also use several other features between the two adjacent tweets in the Markov chain, such as whether they have a same author, whether they constitute a conversation, whether they are in a same thread, whether they contain overlapping emoticons and hashtags, and the similarity of their bag-of-words vectors.

\item Convolutional Neural Network (CNN): a typical neural network for sentiment analysis, which demonstrates competitive performance on sentiment classification~\cite{Kalchbrenner:2014}.

\item Long Short-Term Memory (LSTM): the LSTM approach has been proposed as a state-of-the-art method in the sentiment classification task~\cite{Wang:2015}. It has the ability of capturing complex linguistic phenomena. 
 
\item LSTM-RNN: We replace the second layer of our proposed HLSTM with RNN to assess the influence of the vanishing gradient problem in our task.

\end{itemize}

\begin{table*}
\large
\centering
\begin{tabular}{c|c|c|c|c|c|c|c|c}
 \hline
 \multirow{2}{*}{Method} &
 \multicolumn{3}{c|}{Precision} &
 \multicolumn{3}{c|}{Recall} &
 \multirow{2}{*}{Accuracy} &
  \multirow{2}{*}{Macro-F1} \\
 
 \cline{2-7}
   & Pos & Neu & Neg & Pos & Neu & Neg & \\
   
 \hline
  SVM\(^{multiclass}\) & 0.610 & 0.481 & 0.350 & 0.484 & 0.449 & 0.494 & 0.473 & 0.471  \\

 SVM\(^{HMM}\) & 0.588 & 0.507 & 0.580 & 0.553 & 0.645 & 0.272 & 0.545 & 0.503  \\

 CNN  & 0.281 & 0.570 & 0.529 & 0.650 & 0.569 & 0.443 & 0.568 & 0.503 \\
  
 LSTM  & 0.343 & 0.626 & 0.555 & 0.693 & 0.625 & 0.527 & 0.626 & 0.556 \\
  
 LSTM-RNN & 0.359 & 0.603 & 0.582 & 0.683 & 0.633 & 0.538 & 0.628 & 0.561 \\

 HLSTM & 0.366 & 0.646 & 0.577 & 0.739 & 0.557 & 0.608 & 0.637 & 0.578  \\
 
 HLSTM-f & 0.437 & 0.642 & 0.576 & 0.677 & 0.601 & 0.648 & \textbf{0.641} & \textbf{0.593}  \\
 \hline
 \end{tabular}
\caption{\label{font-table}Experiment results of our model, compared with baselines.}
\end{table*}

\subsection{Experiment Settings}

\noindent\textbf{Implementation}\hspace{5mm} We implement our model by Theano. AdaDelta~\cite{AdaDelta} is adopted to optimize the parameters. We add a dropout layer before the softmax layer to prevent overfitting. The dropout rate is 0.5. We use mini-batch to speed up the convergence. And the batch size is 5, which means we update our parameters after every 5 tweet threads.

\noindent\textbf{Parameter Settings}\hspace{5mm} For CNN, we set the size of convolution filters as 2, 3 and 4 while the number of feature maps is set to 100. The hidden states of the LSTM baseline, the LSTM layer of LSTM-RNN, the first LSTM layer of HLSTM all have a dimension of 128, while the hidden states of the second LSTM layer of HLSTM have 64 dimensions. 

\noindent\textbf{Evaluation Methods}\hspace{5mm} We split our dataset randomly into the training, validation, and test set with a partition of 3:1:1. The validation set is used to tune the hyper-parameters. The classification results are measured by $Macro-F1$ and $Accuracy$.

\noindent\textbf{Word Embeddings}\hspace{5mm} The word embeddings are pre-trained on a large Chinese tweet corpus with 100k tweets, and the embeddings will be fine-tuned during the training process. We set the dimension of the word vector to 128.

\subsection{Results and Analysis}

The experimental results are shown in Table 2. 
We can find that the conventional SVM methods perform relatively poorly. The reason may be due to the fact that SVM features suffer from sparseness and are less effective than word embeddings. SVM\(^{multiclass}\) performs worse than SVM\(^{HMM}\) since SVM\(^{multiclass}\) doesn't take into account any contextual information. And this justifies the effectiveness of utilizing the contextual information.

%CNN and LSTM have both achieved state-of-art performance in some sentiment classification tasks. These two baselines focus on the content of the tweets and don't utilize any context. LSTM overwhelmed CNN by a large margin because LSTM is a sequence model which is capable to leverage the words in a distance of a tweet to generate a more comprehensive representation, so that it can incorporate more semantic information in it. 

Compared with LSTM, our hierarchical LSTM (HLSTM) model achieves much better performance. This result confirms our assumption that utilizing the long-range contextual information is helpful for sentiment classification on such short and noisy texts.

In the LSTM-RNN model, we replace the second LSTM layer in HLSTM with a Recurrent Neural Network layer. The result shows that the LSTM-RNN model performs slightly worse than our HLSTM model. The reason is that RNN suffers from the vanishing gradient problem.

In HLSTM-f, we adopt additional contexts to help our model to encode more contextual information. We use the context features described in Section 3.4. We can see that these additional contexts contribute superior performance, indicating that modeling rich contexts in our model is helpful. However we also notice that HLSTM-f model doesn't have a large promotion over HLSTM in terms of $Accuracy$, and we conjecture that this is because the context features are sparse and most of tweet pairs do not observe such features. 

\begin{table*}
\small
\centering
\begin{tabular}{c|c|c|c}
 \hline
{Tweet Contents in a thread} &
{Gold Standard} &
{LSTM} &
{HLSTM} \\
\hline
\parbox{8cm}{~\\ A: 上午起来，天气晴和。虽有雾霾，雾少霾多。看见太阳，不冷不热。脏了吧唧，搂着云朵。满街破车，来往穿梭。……(When I get up this morning, it's fine and warm and there is haze. I feel neither cold nor hot seeing the sun, which looks very dirty in the haze, surrounded by the clouds. The traffic on the street is heavy. ... ) ~\\} & (-) & (-) & (-) \\
\parbox{8cm}{~\\ --- B: 雾霾是北京人的福利，是北京人自己弄来的美好享受！[哈哈](Haze is the welfare of the people in Beijing. They get themselves a good enjoyment! [smile emoticon]) ~\\} & (-) & (+) & (-) \\
\parbox{8cm}{~\\ ------ A: 东部很多城市都有，石家庄更厉害。(Many cities in the east China has suffered from the haze. It's more severe in Shijiazhuang.) ~\\} & (-) & (+) & (-) \\
\hline

 \end{tabular}
\caption{\label{font-table}An example that HLSTM can classify correctly while LSTM makes an error. Each tweet is associated with three labels, given by the human annotators, LSTM, and HLSTM respectively. }
\end{table*}

\subsection{Case Study}

In this section, we will show some cases to demonstrate the advantages of HLSTM over LSTM. One important case is that the polarities of some sentiment words are changed in some particular context, as exemplified in Table 3. Actually this example is a conversation between two authors. We can find out that in this tweet thread, the HLSTM model predicts correctly all the tweets, while the LSTM model is correct on the root tweet only. By observing the second tweet from author B, we may find that it contains some positive sentiment words such as ``福利 (welfare)'' and ``享受 (enjoyment)''. Then LSTM predicts these words as positive without considering the long-range context. However we can see clearly that the second tweet is negative because it is an obvious sarcasm of the haze in Beijing. In the third tweet, there is an ambiguous sentiment word in Chinese, ``厉害(severe, powverful)''. Without any context LSTM may not be able to decide the correct polarity.

Another common case is that some tweets express agreement or disagreement with the ``ancestor'' tweets. Without context, the LSTM model cannot decide the exact sentiment polarity of the agreeing or disagreeing statement such as ``我完全同意! (I totally agree!)''. However with the long-range context, we can trace back to its ``ancestor'' tweets and judge the polarity more reasonably.

\section{Conclusion}

In this paper, we propose a hierarchical LSTM model with two levels of LSTM networks to model long-range dependency. Moreover, we  adopt rich additional contexts including social context and text-based context to help our model to encode more contextual information. The experimental results has shown that our method is superior to the baselines. Case study further shows that our proposed model can capture the polarity drift with contexts.

\bibliography{acl2016}
\bibliographystyle{acl2016}
\end{CJK}
\end{document}